%% file: main.tex
\documentclass[11pt, a4paper, logo, one column, copyright]{deepmind}

\pdftrailerid{redacted}

\makeatletter
\renewcommand\bibentry[1]{\nocite{#1}{\frenchspacing\@nameuse{BR@r@#1\@extra@b@citeb}}}
\makeatother

\usepackage{kantlipsum, lipsum}
\usepackage{dsfont}
\usepackage{dm-colors}
\usepackage{multicol}
\usepackage{algorithm}
\usepackage{algpseudocode}

\usepackage[authoryear, sort&compress, round]{natbib} 

\graphicspath{{figures/}}

\title{Improving Multimodal Interactive Agents with Reinforcement Learning from Human Feedback}

\reportnumber{} 

\renewcommand{\today}{}

\author[1]{Interactive Agents Team}

\affil[1]{DeepMind}

\input{sections/abstract}

\begin{document}

\maketitle

\input{sections/intro}
\input{sections/approach}
\input{sections/data}
\input{sections/reward_model}
\input{sections/rl}
\input{sections/results}

\input{sections/related}
\input{sections/conclusions}
\input{sections/authors}
\input{sections/acknowledgements}

\bibliographystyle{abbrvnat}
\nobibliography*
\bibliography{template_refs}

\appendix
\renewcommand\thefigure{\thesection.\arabic{figure}}
\setcounter{figure}{0}
\input{sections/appendix}

\end{document}

%% file: sections/abstract.tex
\begin{abstract}
An important goal in artificial intelligence is to create agents that can both interact naturally with humans and learn from their feedback. Here we demonstrate how to use reinforcement learning from human feedback (RLHF) to improve upon simulated, embodied agents trained to a base level of competency with imitation learning. First, we collected data of humans interacting with agents in a simulated 3D world. We then asked annotators to record moments where they believed that agents either progressed toward or regressed from their human-instructed goal. Using this annotation data we leveraged a novel method---which we call ``Inter-temporal Bradley-Terry'' (IBT) modelling---to build a reward model that captures human judgments. Agents trained to optimise rewards delivered from IBT reward models improved with respect to all of our metrics, including subsequent human judgment during live interactions with agents. Altogether our results demonstrate how one can successfully leverage human judgments to improve agent behaviour, allowing us to use reinforcement learning in complex, embodied domains without programmatic reward functions. Videos of agent behaviour may be found at \url{https://youtu.be/v_Z9F2_eKk4}.
\end{abstract}

%% file: sections/intro.tex
\begin{figure}[h]
	\centering
	\vspace{0.25cm}
	\includegraphics[width=.9\textwidth]{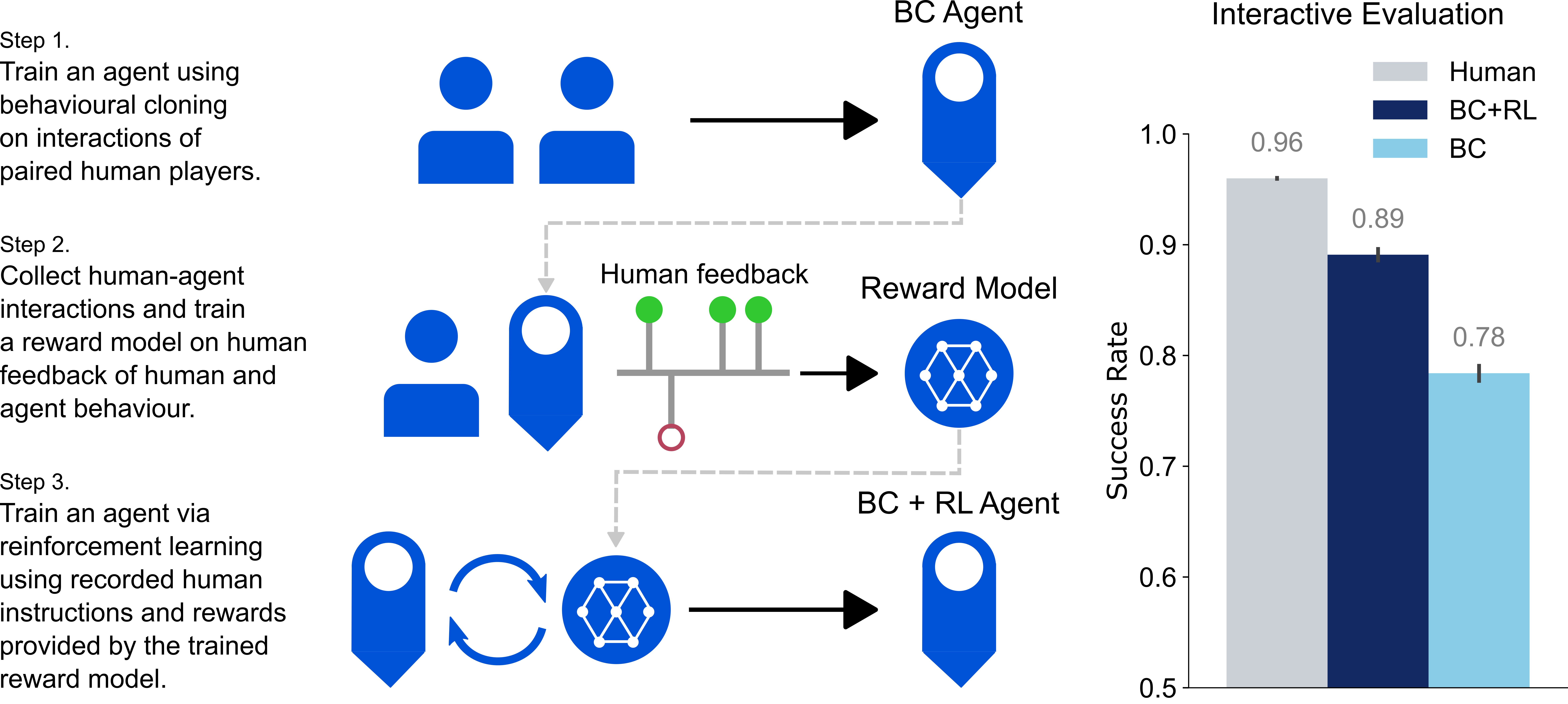}
	\caption{\textbf{Approach.} We started by training an agent via behavioural cloning (BC) from a dataset of human-human interactions, as in \cite{iax2021b}. Then, we collected a dataset of human-agent interactions and, for a subset of episodes, asked humans to provide judgments of progress towards or regression from the human-instructed goal. Next, we trained a network to model this human feedback and obtained a reward model. Finally, we used the reward model to train a new agent that combines behavioural cloning and reinforcement learning from human feedback (BC+RL). \textbf{Interactive evaluation.} We asked humans to interact with agents by providing instructions or questions during 5 min and to evaluate them after each interaction. BC+RL agents obtained a higher success rate compared to BC agents, representing $93\%$ of that achieved by humans in similar conditions.}
	\label{fig:fig1}
\end{figure}

\section{Introduction}

Scalable reinforcement learning (RL) relies on  precise reward functions that are cheap to query.  When RL has been possible to apply, it has led to great achievements, creating AIs that can match extrema in the distribution of human talent \citep{silver2016, vinyals2019}. 
However, such reward functions are not known for many of the open-ended behaviours that people routinely engage in. 
For example, consider an everyday interaction, such as asking someone ``to set a cup down near you.'' For a reward model to adequately assess this interaction, it would need to be robust to the multitude of ways that the request could be made in natural language and the multitude of ways the request could be fulfilled (or not), all while being insensitive to irrelevant factors of variation (the colour of the cup) and ambiguities inherent in language (what is `near'?). 

To instill a broader range of expert-level capabilities with RL, we therefore need a method to produce precise, queryable reward functions that respect the complexity, variability, and ambiguity of human behaviour.
Instead of programming reward functions, one option is to build them using machine learning. Rather than try to anticipate and formally define rewarding events, we can instead ask humans to assess situations and provide supervisory information to learn a reward function. 
For cases where humans can naturally, intuitively, and quickly provide such judgments, RL using such learned reward models can effectively improve agents \citep{ziegler2019, stiennon2020, christiano2017, ibarz2018}.

In previous work we have demonstrated how imitation learning can create agents that capture a diversity of human interactive behaviour. In this work we aim to improve upon our previously built multi-modal interactive agent (MIA, \cite{iax2021b}) by leveraging reinforcement learning grounded in human judgments (sometimes called Reinforcement Learning from Human Feedback, RLHF). 

Our contributions are as follows:
\begin{itemize}
    \item We develop an interface to collect human feedback on two-player interactions in a 3D simulated world.
    \item We introduce the Inter-temporal Bradley-Terry reward model to capture human feedback on individual episodes. 
    \item We show that the reward model improves agent behaviour compared to baselines of imitation-learning agents. These improvements apply both to interactions involving mobile manipulation (environmental locomotion and dexterous physical interaction) as well as tasks involving bi-directional verbal interaction like question-answering. Impressively, in some tested interactions, the agents reach success rates above the average human player.
\end{itemize}

%% file: sections/approach.tex
\section{Approach and main result}

Our general approach is illustrated in Figure \ref{fig:fig1}A. In brief, (1) we trained an imitation-based agent via behavioural cloning (BC) from our dataset of human-human interactions in the Playhouse environment \citep{iax2021b}, also following the cited method. In these human-human interactions, one player (the setter) set tasks for a second player (the solver), who performed the tasks. These tasks involved mobile manipulation and question-answering or combinations thereof. Imitation learning produced an agent that was often competent in the human interactions. (2) We then asked humans to interact with this imitation-trained agent (Section \ref{sec:agent_data}). (3) Human raters annotated a subset of these interactions offline (Section \ref{sec:feedback_data}) by watching videos from the perspective of the solver agent. The raters marked discrete moments where the agent made progress towards or regressed from the goal. (4) We modelled these annotated data, obtaining a ``reward model'' that captures the details of human feedback (Section \ref{sec:reward_model}). (5) Finally, we used reinforcement learning to train the agent to improve with respect to the learned reward model's output (Section \ref{sec:rl}). 

To evaluate our method, we asked humans to interact with agents in real-time and evaluate their successes in each interaction. Agents trained with reinforcement learning \emph{and} behavioral cloning (BC+RL) obtained a higher success rate compared to agents trained via imitation alone (BC), achieving 92\% of performance relative to that achieved by humans in similar positions (Figure \ref{fig:fig1}B).

In the sections that follow, we explain each of these steps and results in further detail.

%% file: sections/data.tex
\section{Collecting behaviour and feedback data}

\subsection{Interactive behaviour} \label{sec:agent_data}

We compiled a dataset of humans interacting with each other in \textit{language games} \citep{iax2021a, iax2021b}. One human played the role of setter, posing tasks and asking questions to a second human playing the role of solver. Interactions were \emph{free-form}, wherein the setter was allowed to choose the tasks/questions at will, or \emph{prompted}, wherein the setter was given a high-level directive to improve around, such as ``Ask the other player to hand you something'' (see Appendix \ref{sec:language_game_prompts}). Episodes lasted 5 minutes and human setters controlled when to give a new instruction based on the human solver's behaviour. 

We then compiled an analogous dataset of humans interacting with agents. In this case, the human always played the role of setter and the agent of solver. For human-agent interactions we used imitation-based agents trained following our previous work \citep{iax2021b}. We also collected human-agent-human interactions, wherein we paired one agent with two humans. As before, one human was the setter and the agent was always the solver. The second human was instructed to watch over the interaction and to take over control of the agent's avatar when the agent was failing at its task (sometimes called `shared autonomy' or `shared control' in the literature \citep{dragan2013policy}). After taking over and helping, the second human handed control back to the agent (see Appendix \ref{sec:shared_autonomy} for details). 

\subsection{Human feedback} \label{sec:feedback_data}

\begin{figure}[t]
	\centering
	\includegraphics[width=1\textwidth]{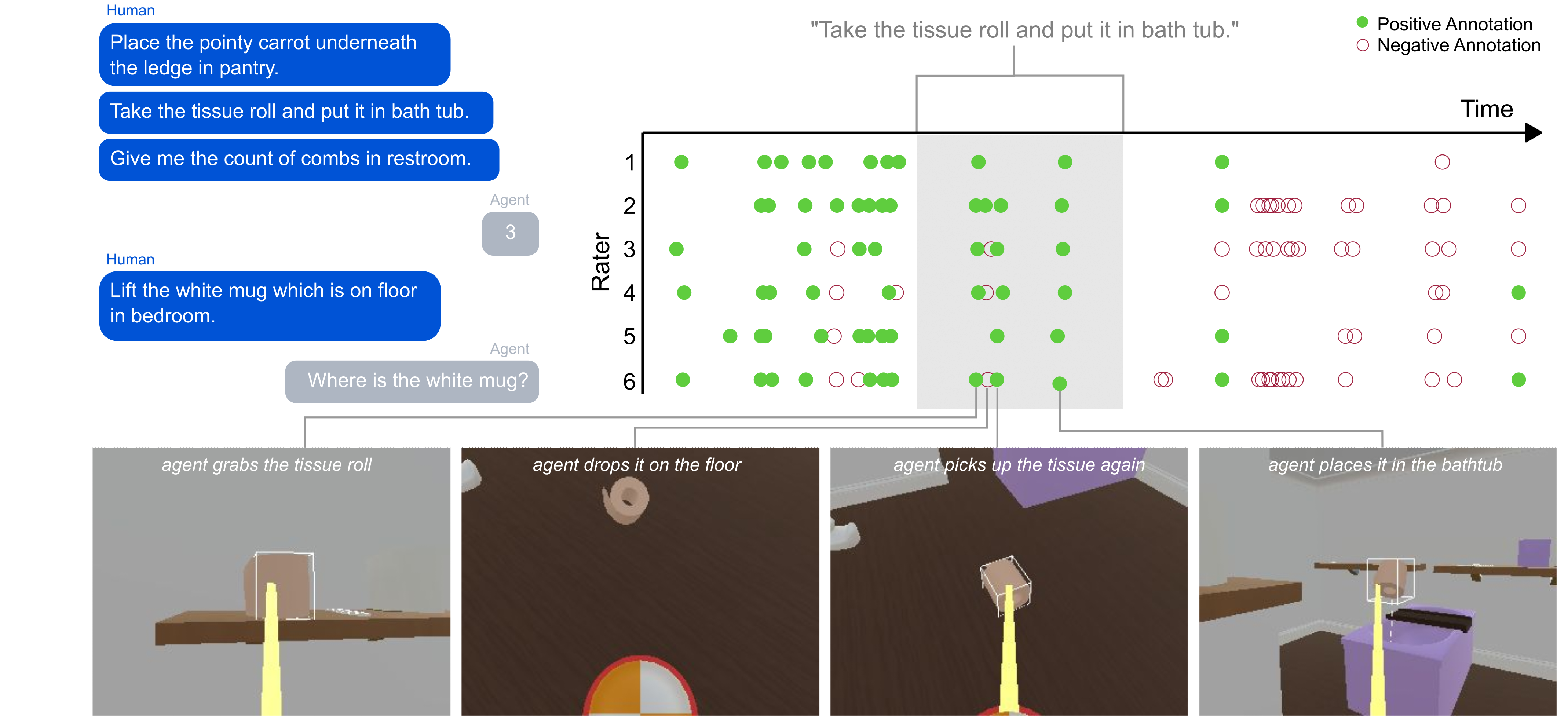}
	\caption{An example episode within the human feedback dataset. The original dialogue between the human and agent is shown on the left, next to a plot of the feedback that 6 human raters gave to this particular episode. Positive and negative feedback marks are represented by filled green and empty red dots, respectively. While 6 ratings were collected for this episode, most of our data was annotated by 1 rater (1.2 on average). A zoomed in segment of the episode (corresponding to the shaded area) is shown at the bottom, together with the feedback of rater $\#$6. During that interaction the human asks the agent to `Take the tissue roll and put it in the bath tub'. Positive and negative marks correspond to moments of progress or regression towards completing the task.}
	\label{fig:fig2}
\end{figure}

To ensure high quality feedback we created an interface where annotators could easily and quickly browse through recorded interactions (including video and dialogue) and mark moments that they believe represented improvement or regression (see Appendix \ref{sec:interface}). This amounted to clicking on discrete time points and providing positive or negative binary annotations. On average, raters provided 14 marks per 5 minute episode.

Figure \ref{fig:fig2} shows an example episode annotated by several raters. As shown in the zoomed-in segment, where the human setter asked the agent solver to ``Take the tissue roll and put it on the bathtub'', raters marked as positive key moments that place the agent closer to solving the task, like picking up the object and placing it in the correct place. They also placed negative marks at times of regression, like dropping the object on the floor.

In addition to collecting feedback on human-agent interactions, we also asked for human feedback on human-human and human-agent-human interactions. Human-human interactions typically demonstrate long sequences of desirable behaviour, and human-agent-human interactions typically demonstrate mistakes in agent behaviour and associated corrections of the mistakes. In total, we collected 5,104,000 individual feedback marks over 364,690 episodes.

%% file: sections/reward_model.tex
\section{Training reward models from human feedback} \label{sec:reward_model}

In much past work on preference elicitation, humans express a preference between two options \citep{akrour2012, akrour2014, christiano2017, ibarz2018, stiennon2020}. It is assumed that they choose option $A$ over option $B$ if they assign higher utility to it: $U(A) > U(B)$ \citep{bradley1952}. This works well if humans can easily identify whether $U(A) > U(B)$, which tends to be the case when humans can intuitively compare $A$ and $B$. Following the old saw, it is easier to compare two apples than an apple and an orange: very dissimilar experiences may be hard to compare. From experience in our setup, two fully open-ended interactions---each with their own Playhouse setting, stream of dialogue, and tasks---are quite dissimilar (an ``apples vs. oranges" scenario), making comparisons difficult.

One option to overcome these difficulties is to tightly control the initial conditions of the environment and ask humans to attempt to have the same interaction twice---with the agent performing its part differently each time, providing information for evaluation. However, this method can be hard to put into practice in many contexts, for example robotics, where environmental conditions are hard to control; furthermore, for multi-turn or long interactions, even shared initial conditions can lead to divergent, incomparable trajectories.

Given these considerations, instead of asking humans to express preferences over entirely different interactions, we asked humans to express preferences between different moments within the same interaction. In practice, we specifically asked humans to judge whether if they had observed events indicating conspicuous progress toward the current instructed goal or conspicuous errors or mistakes. We then drew a correspondence between these positive and negative events and positive and negative preferences. More precisely, if a rater positively marks time point $t_2$ following time point $t_1$, then we consider this to reflect the opinion that the rater prefers time $t_2$ to $t_1$. Because these judgments take place across time, we call them `inter-temporal' judgments in keeping with the nomenclature in economics on `inter-temporal choice' \citep{lewis2008}.

We built a network to model the utility of a trajectory $U_{\theta}(x_{\leq t})$ and trained it to classify pairs of sub-trajectories of the same episode, according to the sign of the human annotation in between, 
\begin{equation}
    L(\theta) = \mathbb{E}_{\text{D}}[ \textbf{prefer}(x_{\leq{t}}, x_{\leq{t'}}) \ln \sigma(U_{\theta}(x_{\leq{t}}) - U_{\theta}(x_{\leq{t'}}))]
\end{equation}
where $x_{\leq t} = (x_0, ..., x_t)$ is the sequence of observations (images and dialogue) up to time $t$, $t$ and $t'$ are two time points of the same trajectory, $D$ is the dataset of human feedback annotations and $\textbf{prefer}(x_{\leq{t}}, x_{\leq{t'}})$ is an indicator function that is $1$ if $t'$ followed $t$ and was marked positive or if $t$ followed $t'$ and was marked negative. We trained the model on pairs of time points that had one or more annotation marks of the same sign in between them and we ignored pairs of time points that had multiple annotations of different signs in between. More details on the Inter-temporal Bradley-Terry model may be found in Appendix \ref{sec:ibt_method}

The reward model network consisted of the same components as the agent: a series of ResNet blocks processed the incoming image, while language was tokenized and embedded. A multi-modal Transformer combined these inputs and its output was fed to an LSTM memory. The output of the LSTM was passed through an MLP with one output, which represented the utility. More details on the architecture are included in Appendix \ref{sec:architecture}.

In addition to training on human feedback annotations, we found that the reward model benefited from training with two additional objectives: behavioural cloning on human-human data (BC) and contrastive self-supervised representation learning (CSS). We augmented the reward model architecture with the same policy used in the agent and used behavioral cloning as an auxiliary loss. To improve generalization we used the same cross-modal self-supervised loss as in the agent \citep{iax2021b}. Given that the reward model shared the same architecture as the agent, we initialized its parameters with that of a pretrained BC agent and combined the IBT, BC and CSS losses with equal weights.

Finally, during reward model training we augmented our dataset by creating artificial examples of incorrect behaviour. For each episode of the dataset, with a probability of .33, we generated a negative example by either (1) randomly swapping the setter's instruction with another one in a training batch, which almost always rendered the trajectory of solver behavior inconsistent with said instruction, or by (2) randomly shuffling the solver's language output with another utterance in the batch, which similarly led to more examples of incorrect dialogue.

As an alternative to the IBT model, we also considered modeling human feedback by directly building a generative model of the annotation data. We replaced the IBT reward model head by an autoregressive (AR) module, while keeping all the other components equal. The autoregressive head is a three-way classifier which modelled, for each time step, the probability of a positive, negative or no annotation. Preliminary experiments with the AR reward model showed weaker results compared to the IBT model, see Appendix \ref{sec:autoregressive} for details.

%% file: sections/rl.tex
\section{Reinforcement learning with IBT-based reward models} \label{sec:rl}

Training agents using RL during real-time interactions with humans is difficult given the stability and data efficiency of most existing algorithms. An alternative option is to collect vast amounts of human-agent interactions and then use offline RL. Or, one could use a variety of model-based control to train. In our context, one could train a model to imitate the human setter and use online RL during simulated interactions between two agents (see Appendix \ref{sec:pretrained_setters} for a first step in this direction).

We primarily explored a third option, which we call `setter-replay'. Setter-replay involves drawing an episode from our database of interactions, recreating the same Playhouse initial configuration and replaying the human setter behaviour step-by-step while allowing agents to act freely. This approach has important limitations. First, it does not easily permit learning back-and-forth interactions: because the setter is replayed from data, it cannot react to the agent's actual behaviour (e.g., to answer a clarifying question). Second, the set of interactions is limited to the data that we have collected. Nevertheless, we found that setter-replay was a useful starting point and enabled us to make improved agents as judged by humans.

As baseline, we used the same agent as \cite{iax2021b}. This agent was trained by BC and cross-modal self-supervised learning. For the BC+RL agent, we used the same architecture, except for an additional value head (an MLP that receives the same inputs as the policy heads). We initialized the agent with the parameters of a pretrained BC agent and continued to train it with a combination of BC and RL.

We used RL to optimize the relative change in utility during the episode $U(x_T)-U(x_0)$. Given this specification, the per-step reward was given by
\begin{equation}
    r_t \stackrel{\text{def}}{=} U(x_{t+1}) - U(x_t),
\end{equation}
where the utility $U(x_t)$ at each time step was provided by the IBT reward model. We adopted the same distributed training setup as in \cite{iax2021a}, using the Importance Weighted Actor-Learner Architecture \citep{espeholt2018} for RL. We weighted this with a co-trained BC loss on batches of human-human trajectories. Thus, we refer to the whole optimisation setup as `BC+RL'. See Appendix \ref{sec:hypers} for more details.

%% file: sections/results.tex
\section{Results}

\subsection{Reward model}

\begin{figure}[t]
	\centering
	\includegraphics[width=\textwidth]{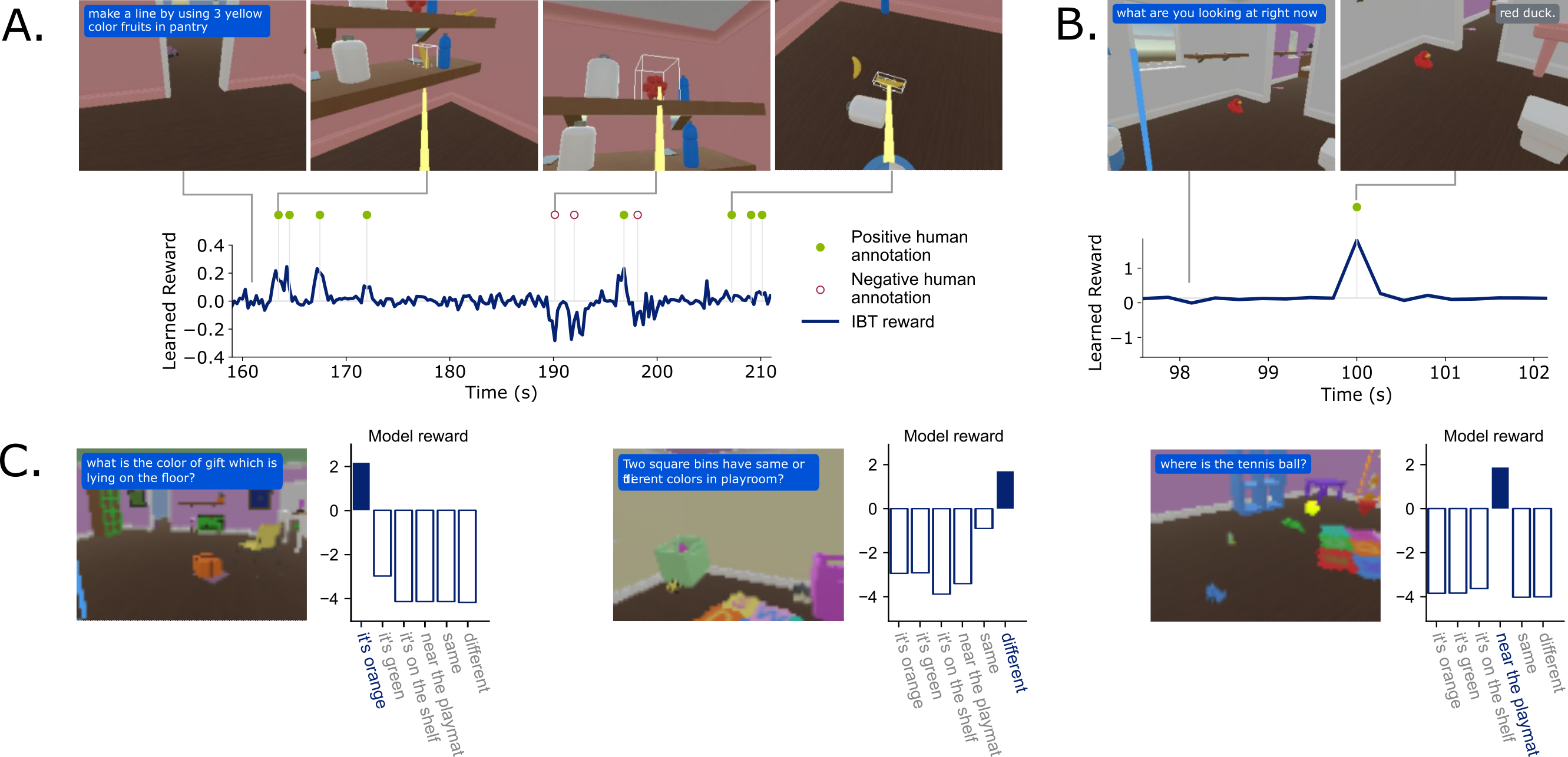}
	\caption{\textbf{A, B} The temporal differences $r_t = U(x_{t+1})-U(x_t)$ of the reward model traces for a human-agent episode of the validation set. The blue trace is the reward assigned by the reward model at each time step. The green (filled) and red (empty) dots are the positive and negative annotations, respectively, marked by humans. The reward model successfully rewards key moments in the episode matching the annotations given by humans. \textbf{C} Reward produced by the reward model to different answers for the same question. The model correctly gives positive rewards to the correct answers and negative rewards to wrong ones. Moreover, the reward is less negative for incorrect but plausible answers than for incoherent answers. }
	\label{fig:fig4}
\end{figure}

Inter-temporal Bradley-Terry reward models qualitatively capture human feedback. Figure \ref{fig:fig4}A,B shows examples of the trained reward model scoring episodes from the validation set (see an example video at \url{https://youtu.be/PvVOjNoW5oA}). The blue trace is the reward assigned to each time step by the reward model and the filled green and empty red dots indicate the ground-truth positive and negative annotations marked by humans, respectively. In Figure \ref{fig:fig4}A, the human asked the agent to make a row of yellow fruits. The reward produced by the reward model increases when the agent picks up a yellow fruit and when it places it on the floor, which coincides with positive human annotations. Conversely, the reward decreases when the agent incorrectly grabs a red fruit, which also aligns with a negative human annotation. 

Reward models can also score language utterances from the agent. Figure \ref{fig:fig4}B shows the output of the reward model to the agent answering a question posed by the human. To confirm the reward model's ability to score language outputs, in Figure \ref{fig:fig4}C we re-ran the reward model on the same episode replacing the correct answers by a set of incorrect ones. The reward model correctly gives positive rewards to correct answers and negative rewards to wrong answers. Moreover, the reward is less negative for incorrect but plausible answers than for incoherent answers.

\subsection{Modelled reward optimisation}

\begin{figure}[t]
	\centering
	\includegraphics[width=0.5\textwidth]{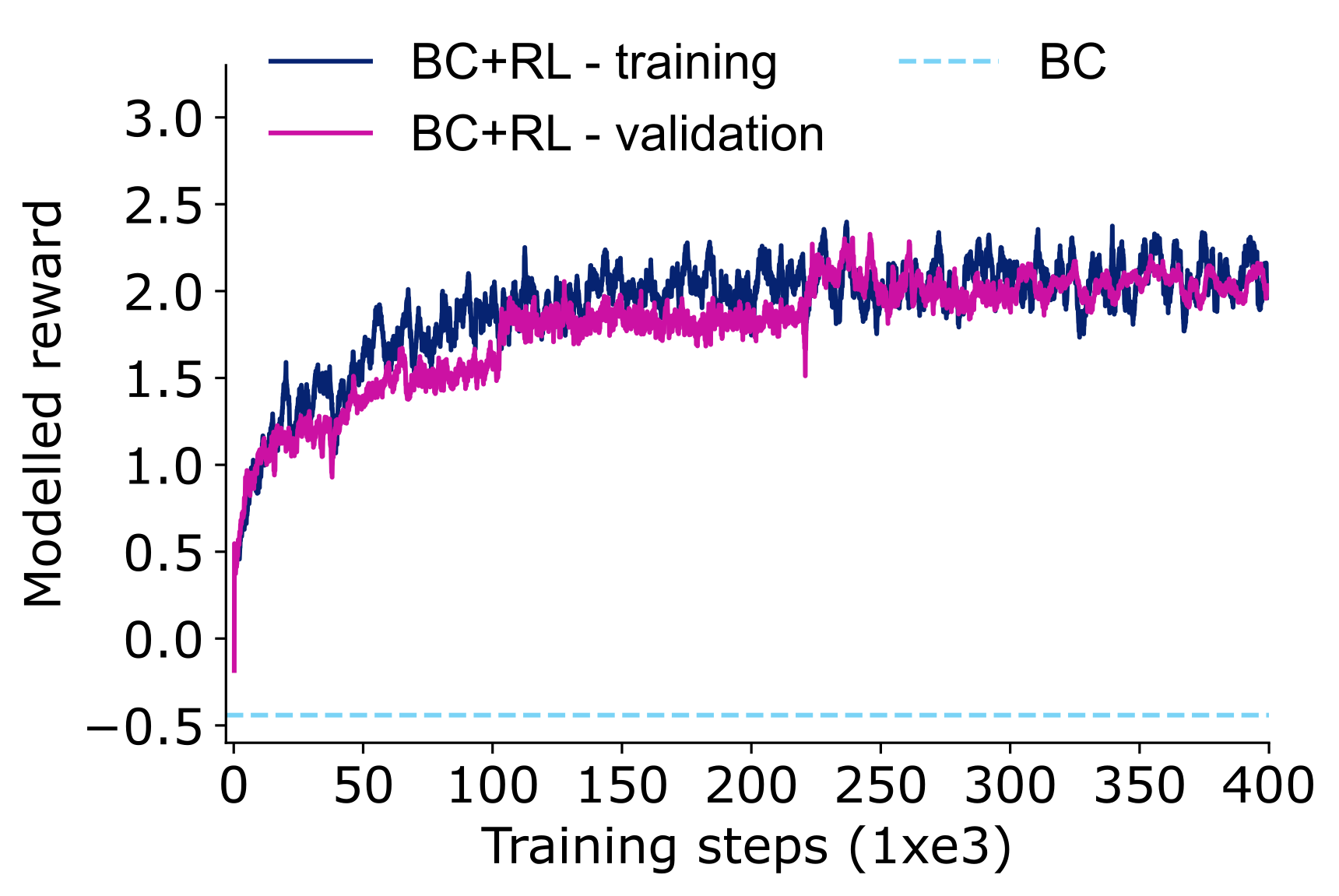}
	\caption{Averaged modelled reward per episode as a function of training steps (learner updates) for data of the training and validation set. Agents trained by RL optimise the modelled reward during training and this is conserved on episodes not used in training. For reference, we plot the reward obtained by the BC agent which we use for initializing the weights of the BC+RL agent.}
	\label{fig:fig5}
\end{figure}

Agents trained with BC+RL optimised the modelled reward over the course of training. Figure \ref{fig:fig5} shows the average per episode modelled reward as a function of training time (learner updates). The initial value of accumulated reward matches that of the BC agent, as we initialized the BC+RL agent with a pretrained BC agent. During the course of training the accumulated reward of the BC+RL agent monotonically increases. Importantly, the accumulated reward is similar for training and held-out episodes, indicating that agents are not overfitting to the particular episodes and instructions used for training.

\subsection{Scripted Probe Tasks}

\begin{figure}[t]
	\centering
	\includegraphics[width=\textwidth]{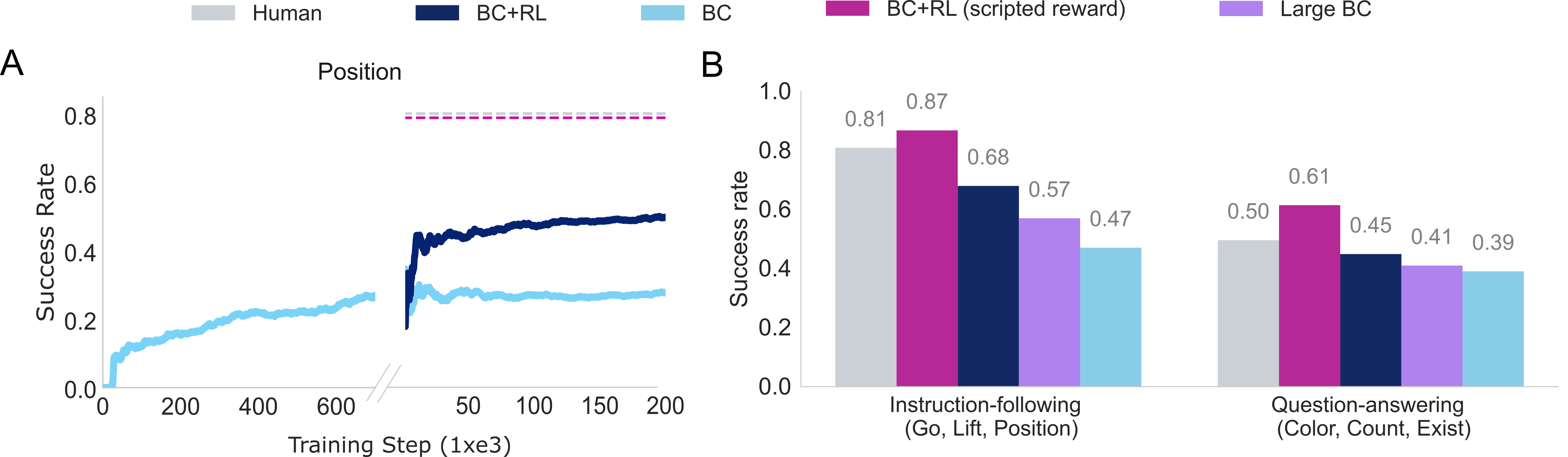}
	\caption{Scripted probe tasks. \textbf{A} Success rate of BC and BC+RL agents on the "Position" held out scripted probe task as a function of training steps (model updates). This task, which is only used for evaluation, supplies a pre-scripted instruction of the form "Put object <X> next to <Y>" where <X> and <Y> are procedurally generated object names present in the Playhouse. After an initial phase of BC pre-training, we compare an agent that keeps being trained with the BC objective with one that additionally optimizes the modelled reward by RL. The BC+RL agent attains approximately twice the success rate compared to the BC agent. \textbf{B} Success rate on six scripted evaluation tasks. Note that while models trained with scripted rewards perform well on scripted probe tasks (BC+RL (scripted reward)), they struggle when interacting with humans (see figure \ref{fig:fig7}), indicating the limits of using scripted rewards when the ultimate goal is naturalistic interactions with humans.}
	\label{fig:fig6}
\end{figure}

As in \cite{iax2021a, iax2021b}, we can use scripted probe tasks as simple, automatic evaluation methods. These tasks provide procedurally generated instructions in synthetic language and a hard-coded function to detect task completion. These evaluation tasks measure skills like counting, identifying colors, lifting objects and positioning one object next to another. Crucially, these tasks were used for evaluation only; agents never received rewards from the environment for the purposes of optimization.

Figure \ref{fig:fig6}A shows the performance of the BC vs BC+RL agent on the `Position' task as a function of training time. In this task, a pre-scripted instruction of the form "Put object <X> next to <Y>" was provided to the agent, where <X> and <Y> are procedurally generated names of objects present in the current configuration of the Playhouse. During BC pre-training, the agent improves at the task. After approximately 700K model updates, performance plateaus and continuing training with BC does not improve the agent's success rate as defined by the task's scripted reward. Adding RL training on the setter-replay environment significantly improved agent performance, achieving approximately double the success rate of the BC agent.

Figure \ref{fig:fig6}B shows the final score on six scripted probe tasks (Go, Lift, Position, Color, Count and Exist; see details in \cite{iax2021a}). BC+RL agents perform better than BC across all tasks and even improve over larger BC models, with more than double the amount of parameters (71M vs 165M). As reference, we include the performance of humans and of BC+RL agents that are trained on these specific tasks with scripted ground-truth rewards. Note that while models trained with scripted rewards perform well on scripted probe tasks (BC+RL (scripted reward)), they struggle when interacting with humans (see Figure \ref{fig:fig7}), indicating the limits of using scripted rewards when the ultimate goal is naturalistic interactions with humans.

\subsection{Standardised Test Suite (STS)}

\begin{figure}[t]
	\centering
	\includegraphics[width=\textwidth]{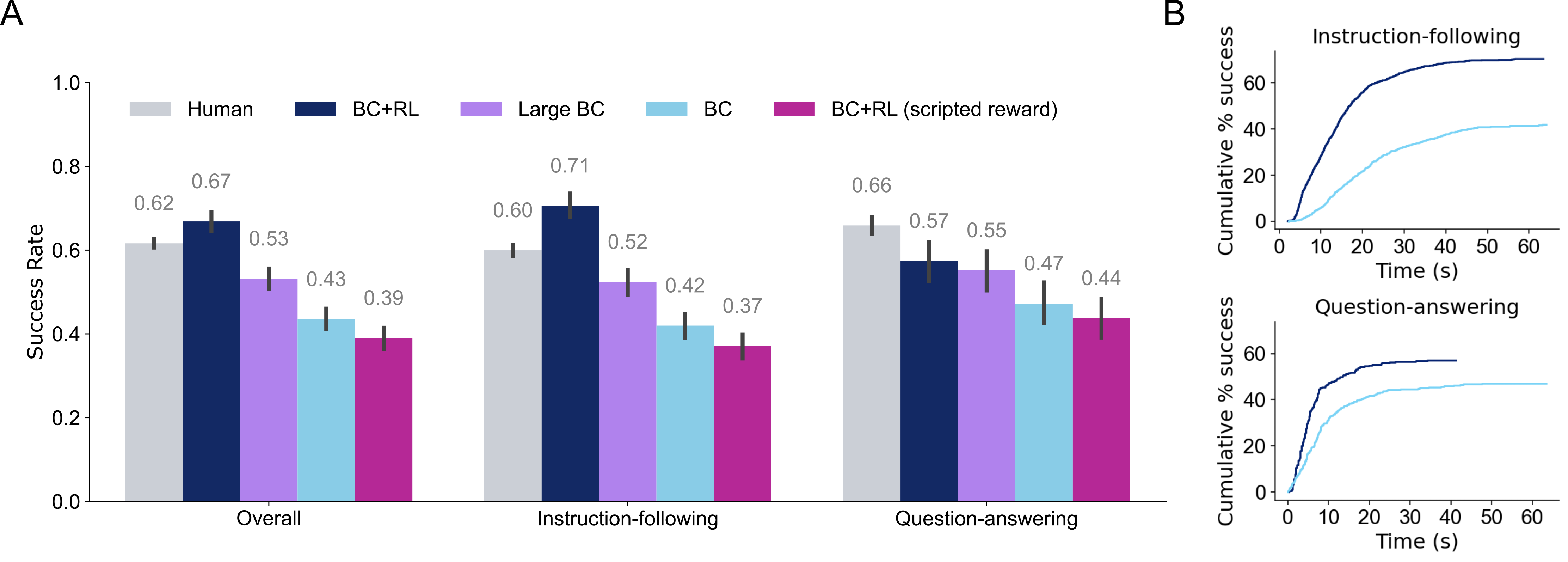}
	\caption{Standardised Test Suite evaluation. \textbf{A} We adopt the method from \cite{iax2022} to evaluate our agents. We collected a set of 162 scenarios, extracted from human interactions, and generated behavioral continuations for humans to evaluate. The plot shows the averaged success rate across all scenarios and split into "Instruction-following" and "Question-answering". Agents trained with BC+RL are scored by humans significantly better than agents trained by BC. The BC+RL agent also performs better than a BC agent with more than twice the number of parameters (71M vs 165M). On the other hand, agents trained by RL with ground-truth reward on the six scripted tasks score poorly, presumably because they cannot generalize to the variety of human interactions in these scenarios. \textbf{B} Cumulative percentage of success as a function of episode length. We measure how many scenarios were successfully completed as a function of the time it took to complete them. Agents trained with RL not only complete more scenarios successfully, but also require less time to achieve success.}
	\label{fig:fig7}
\end{figure}

Scripted probe tasks are useful as a quick and automatic method to evaluate agents, but have important limitations. They require significant effort to create and the hard-coded detection of task completion is brittle \citep{iax2022}. Additionally, overall performance on these tasks does not necessarily correlate well with human evaluation of natural interactions. To overcome these limitations we use the Standardised Test Suite, an evaluation method developed in prior work \citep{iax2022} which is fast, precise and correlates well with human judgement during online interactive evaluation. We collected a set of 162 scenarios (see Appendix \ref{sec:sts_scenarios}) randomly extracted from human-agent interactions. Scenarios were selected from episodes where humans were instructed to challenge the agent, i.e. produce instructions or questions which they thought the agent would struggle with. For each scenario we generated 10 behavioral continuations per agent and sent these episodes for humans to evaluate as success or failure. 

Figure \ref{fig:fig7}A shows the average success rate across all scenarios for BC versus BC+RL agents as well as split into `Instruction-following' and `Question-answering'. Agents trained with BC+RL are scored by humans significantly better than agents trained only by BC. BC+RL agents also perform better than larger BC agents with more than twice the amount of parameters (71M vs 165M). On the other hand, agents trained via RL on the six scripted probe tasks using scripted ground-truth rewards score poorly, presumably because they cannot generalize to the variety of human interactions in these scenarios.

Training with RL leads to more skilled and efficient agents. In Figure \ref{fig:fig7}B we show the cumulative percentage of scenarios that were successfully completed as a function of the time it took to complete them. Agents trained with BC+RL not only successfully complete more scenarios but also require less time to achieve success. (See example videos at \url{https://youtu.be/A0BVULc7aUk}.)

\subsection{Interactive Evaluation}

\begin{figure}[t]
	\centering
	\includegraphics[width=\textwidth]{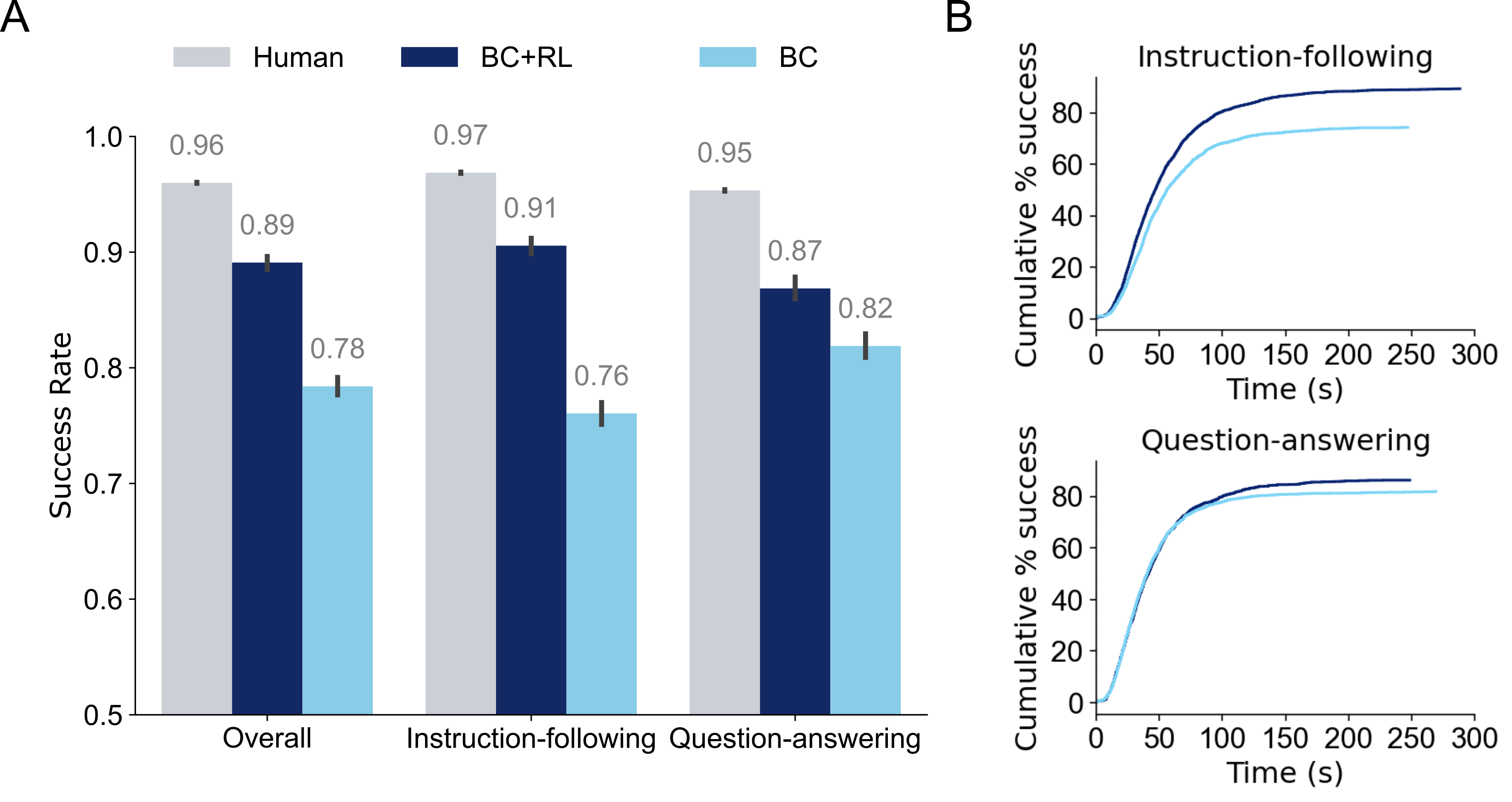}
	\caption{Interactive evaluation. \textbf{A} We asked raters to evaluate our agents in online real-time interactions. Humans played the role of setters, providing instructions or questions during 5 min and judging the agents' success. We find that BC+RL agents obtain higher success rate compared to BC agents, achieving 92$\%$ the performance of humans in similar conditions. \textbf{B} Cumulative percentage of success as a function of time. During interactive evaluation we measure the time to completion as the amount of time elapsed between the instruction or question provided by the human and the moment when human judged that the task was completed. Agents trained with BC+RL succeed faster than BC agents, particularly for instruction-following tasks.}
	\label{fig:fig8}
\end{figure}

Our aim is to improve the ability of agents to interact naturally with humans. Therefore, the ultimate metric we care about is how humans judge live interactions with agents. So, upon training agents, we paired humans setters with agent solvers for episodes of 5 minutes. Human setters provided an average of 3.5 instructions per episode and, after each interaction, reported binary feedback as to whether the interaction was successful or not, which we report as success rate.

The BC+RL agent achieves an overall success rate of 89$\%$, representing 93$\%$ of that achieved by humans in similar positions, significantly more than the agent trained only by BC (Figure \ref{fig:fig8}A). Moreover, the time it takes the BC+RL agent to complete each interaction is lower than that of the BC agent, as indicated by the cumulative number of success episodes as a function of time (Figure \ref{fig:fig8}B). 

\subsection{Model scaling} \label{sec:scaing}

\begin{figure}[t]
	\centering
	\includegraphics[width=\textwidth]{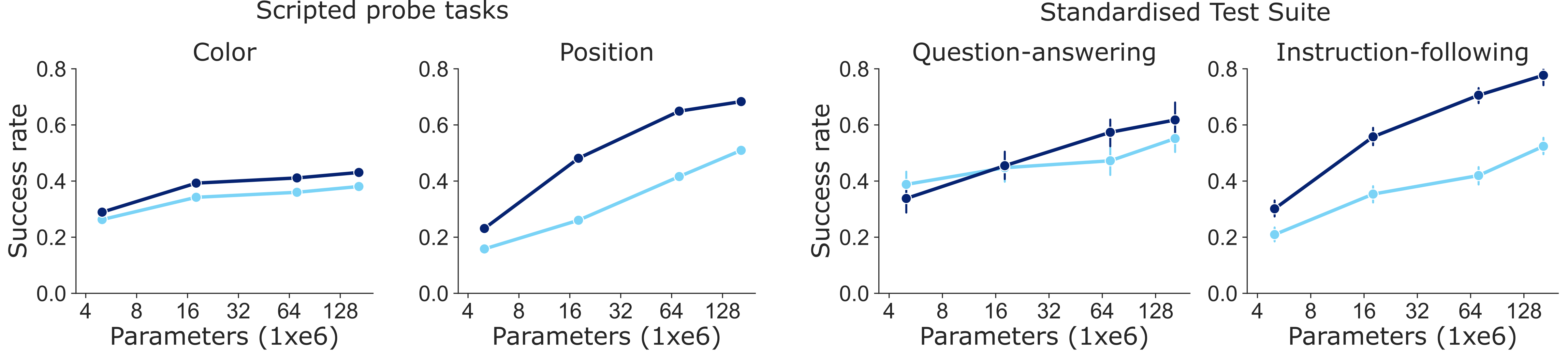}
	\caption{Scaling performance of the BC+RL and BC agents. For smaller models, we scaled the depth of the vision encoder and the embedding size of the Transformer by 0.25 and 0.5, obtaining models with 5M and 18M parameters, respectively. To obtain a larger model, we scaled the depth of the vision enconder's ResNet by 2 and the Transformer's embedding size by 1.5 times, resulting in a model with 165M parameters. We used these architectures to train the BC model, the reward model and the RL agent. Both in scripted probe tasks and STS evaluation, the BC+RL agents outperforms the BC one, across all scales and show monotonic increase in performance as a function of model size.}
	\label{fig:fig9}
\end{figure}

It is well known that larger models improve supervised learning in a predictable, law-like manner as long as the quantities of data are sufficient. This is a general phenomenon and has been studied in language \citep{kaplan2020}, vision \citep{dosovitskiy2020}, and other prediction problems \citep{reed2022} as well as offline RL \citep{lee2022}. However, such scaling phenomena have not been easily observed in online RL except in narrow contexts \citep{neumann2022}. (This is partially because many common RL environments have very little diversity.) 
The relevance of these scaling studies is to understand whether and how much changes to model size would improve performance.
In our context, with a heterogeneous architecture, a heterogeneous loss function of BC+RL, a pretrained reward model that provides reward to an online RL algorithm, and the dependent variable of binary success or failure on tasks, it is by no means obvious that scaling with model size should be observable --- and, if observable, law-like. Yet examining it empirically is still important to determine if performance has saturated with model scale.

To investigate scaling performance, we tested the performance of models with different sizes (5M, 18M, 71M (baseline) and 165M) by scaling the vision encoder's ResNet depth and the embedding size of the Transformer and LSTM. We applied the same model scaling to all components of training: BC pre-training, reward model and RL agent. Figure \ref{fig:fig9} summarizes the scaling results. To understand the effect of scaling on question-answering and instruction-following tasks, we evaluated models on two scripted probe tasks: ``Color'' (What is the color of <X>?) and ``Position'' (Put <X> next to <Y>). Compared to the BC baseline, BC+RL achieves better performance across all scales. More importantly, we see a monotonic increase in performance with large models, suggesting BC+RL also benefits from model scaling, and is not saturated. Over all, our results demonstrate that, for a given model size, BC+BL can improve performance, and it appears the benefits are more prominent in instruction-following tasks (locomotion and dexterous manipulation) than in question-answering tasks. 

\subsection{Iterative improvement}

\begin{figure}[h]
	\centering
	\includegraphics[width=.9\textwidth]{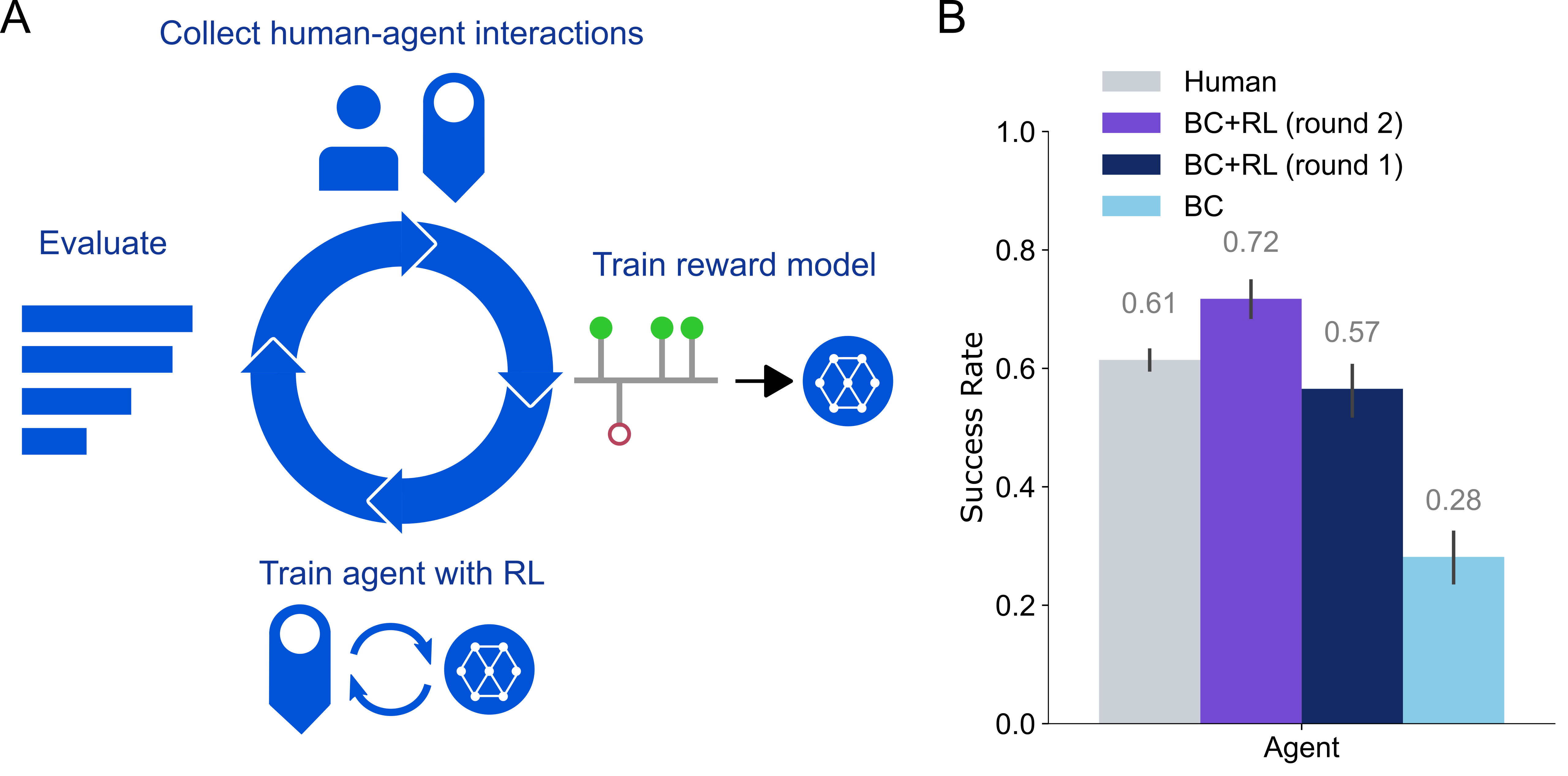}
	\caption{Iterative improvement. \textbf{A} In each iteration, we (1) deploy an agent to interact with humans and collect feedback on its behavior, (2) update the reward model by training on this new data, (3) train a new RL agent using the updated reward model, (4) evaluate and repeat. \textbf{B.} We created a specific STS for this task, with episodes taken from the validation set. Each iteration of iterative improvement produces a better agent and by the second iteration the agent scores better than humans on this task. }
	\label{fig:fig10}
\end{figure}

Having demonstrated the benefits of training agents by RL with reward models learned from human feedback, we finally studied whether we can iterate this process to continually improve an agent. The idea behind iterative improvement (Figure \ref{fig:fig10}) is to run a cyclic process where we (1) deploy an agent to interact with humans and collect feedback on its behavior, (2) update the reward model by training on this new data, (3) train a new agent by RL with the updated reward model, and (4) evaluate and repeat.

As a proof-of-concept, we focused on a particular task (`Build a tower') and ran two iterations of this cycle. We started by deploying the BC agent and collected 55K episodes where human setters are prompted to `Ask the agent to build a tower' with some specialisation. Humans come up with diverse instructions, e.g. building towers of particular objects, in specified locations, etc. We annotated these episodes with human feedback and trained a reward model, as explained in Section \ref{sec:reward_model}. This reward model was then used to train a first BC+RL agent (round 1). In the second iteration, we deployed this BC+RL (round 1) agent and collected another 66K episodes and annotated them. We updated the reward model by training on this data appended to the one from the previous iteration and used it to train a new BC+RL (round 2) agent. Both the reward model and BC+RL agent were initialized from the weights of the pretrained BC agent. To evaluate agents we created a specific standardised test suite for this task, with episodes taken from the validation set. Figure \ref{fig:fig10} shows that each iteration of this cycle produces a better agent. Moreover, by the second iteration the agent scores better than humans on this task. 

While we demonstrated that we can iterate the RL process to repeatedly improve agent behavior, this is only a first step: We have restricted the set of interactions to a single task and focused on an instruction-following skill (`Build a tower'). Preliminary experiments on a question-answering task (`Name  objects'), which may relate to our generally stronger improves from RL on instruction-following than question-answering discussed in \ref{sec:scaing}. Moreover, further research is needed to understand how performance depends on the amount of data collected --- and how best to interleave training and interaction.

%% file: sections/related.tex
\section{Related Work \& Discussion}

Psychologists and economists have used choice preferences to estimate utility since the middle of the 20th century \citep{luce1965preference}. 
But for reasons of practicality, almost all algorithmic development in the field of RL has used simple, programmed reward functions.
This approach is entirely sensible since potential alternatives---measuring human preferences, for example---can be time-consuming, expensive, and difficult to interface with algorithmic experiments on a computer.
However, this also means that reinforcement learning is generally untested on tasks whose solutions are qualitative and require human judgements.

On the other hand, in supervised learning the need to incorporate human judgements has long been taken for granted. Machine learning practitioners routinely---though not exclusively \citep{dsprites17, nikolenko2021}---collect human annotations or labels for image, text, audio, and video data \citep{mnist,imagenet}. These human-derived labels are used to create input/output $(x,y)$ data sets that can be fixed, enabling other parts of the machine learning pipeline, such as network architectures, learning rules, and regularization schemes, to be varied. Working with human annotations can be easier in the supervised learning setting since these annotations only need to be collected once to enable many downstream experiments.

In reinforcement learning, human judgements used in lieu of programmatic reward functions are most naturally collected for a particular policy or set of policies.
But this introduces a variety of difficulties --- namely, human preferences are estimated only for observed behaviours, rather than for all relevant behaviours. 
As a consequence, most human-derived annotation processes in RL, ours included, are expensive, dynamic, and \emph{local}: a reward model is constructed that can only offer useful judgments in the neighbourhood of the current policy; more annotation must be collected as reinforcement learning training proceeds.

Early experiments that used human annotations in place of programmatic reward used simple domains \citep{akrour2011,akrour2012,akrour2014,el2016}. There has been a recent surge of work in more complex settings, bringing forth the techniques into the realm of deep RL \citep{christiano2017, ibarz2018}, large language modeling \citep{ziegler2019, stiennon2020, thoppilan2022l, bai2022, ouyang2022}, and robotics \citep{cabi2019}.

Much of this recent work leverages human \emph{preferences} by asking humans to compare two task solutions side-by-side (two outputs from the same AI agent, outputs from different agents, or from a human and an agent, and so on), and indicate which they prefer. Preference models are created from these data, and are used to administer scalar rewards to agents trained with conventional RL techniques. While this basic recipe works well with language modeling, it may not be the best form of feedback for more complex, real-time, multi-modal, interactive data, where it is more difficult to elicit comparable solutions \citep{wirth2017}. In the context of e.g. text summarisation, language models can easily be prompted with exactly the same text and then given the opportunity to produce two responses, which may be compared easily. However, when environmental conditions cannot be \emph{reset} so easily, it is harder to construct a simple comparison. For example, in robotics it may be difficult or tedious to reset the system after the robot has already manipulated the environment. Consequently, preference estimation requiring nearly identical, side-by-side judgements has a somewhat restricted scope of application.

An alternative form of feedback is single trajectory judgment (a.k.a. evaluative feedback \citep{zhang2021}) wherein humans watch agents behave and provide a scalar signal representing their judgement of how good or bad that behaviour was. This type of feedback is usually given on state-action pairs (as opposed to trajectories), requiring more human effort. The main challenge with this feedback is how to interpret it in the context of RL. Previous work has treated evaluative feedback as reward \citep{pilarski2011}, value \citep{knox2009} or advantage \citep{macglashan2017}. As opposed to learning from preferences, scalar feedback in single trajectories is subject to arbitrary scale choices, reward shaping and/or reward engineering \cite{zhang2021}. 

Our work can be seen as a bridge between preference and single-trajectory feedback. We collect discrete binary feedback on single trajectories and leave the choice of which time points to judge to the humans. We do not try to compare two unique scenarios that we arrange to be as similar as possible; yet, we still interpret feedback as a \emph{comparison} or preference between sub-trajectories. From this we extract a signal that at each time point indicates human preferences over agent behaviour.

There remain many questions about how to improve on our paradigm for creating goal-directed agents that interact with humans. Empirically, our performance gains for movement tasks are significantly larger than for language-output tasks. On this front, there may be ideas that can be borrowed from recent studies on reinforcement learning for turn-based dialogue agents \citep{bai2022,glaese2022improving}. Another potential limitation of our approach (and the recent turn-based dialogue studies) is that we may struggle to optimize \emph{long-horizon preferences}. Humans' annotation of improvement/regression may tend to focus on local-in-time judgements that are easier to make, and our annotation approach treats each improvement/regression indicator as equally important. In this sense, the method relies on humans to be able to judge intermediate progress and regression. If humans fail to score intermediate progress or regression correctly, the resulting trained reward models may fail to take into account long-term consequences, which will create myopic agents. Such agents may be hesitant to take an action which has frequently been marked as a regression, even if it is the only way to move towards a longer term goal requested by the human. The obvious solution to this limitation is to capture and model human judgements over long interactions, though this induces potential issues with scarcity of reward feedback. 

%% file: sections/conclusions.tex
\section{Conclusion}

In this paper we have developed a framework for building multi-modal grounded language agents that can solve complex, ambiguous, human-instructed tasks and master them using reinforcement learning against human judgments. The framework presented in this work is relatively general and could be deployed without dramatic modification to develop digital assistants, video game AIs, and language-instructable robots. To develop the framework, we have made progress in human preference reward modeling by eschewing comparison of different trajectories; instead, we compare the same trajectory across time, which enables the framework to apply to any time series that humans can judge for goal-directed behaviour --- even possibly real-world videos. By applying the framework to initial policies with behavioural priors, we have shown that the performance of the agents improves substantially, and, according to narrow diagnostics, is better than average human performance.

While our framework is broadly useful, we recognise several possible paths to improvement. One clear direction for future work is to combine our approach with large self-supervised models trained on internet-scale data \citep{brown2020, radford2021, alayrac2022, ramesh2021, jia2021}. We expect this may enable us to build broader behavioural priors upon which to apply reinforcement learning, which may in turn significantly reduce the amount of data required to produce goal-directed agents. 

We hope that this work along side developments in large-scale modelling will inspire work on agents that can interact with humans in rich, complex, sensorimotor worlds.

%% file: sections/authors.tex
\newpage
\section{Authors \& Contributions}

\textbf{Josh Abramson} contributed to agent development, imitation learning and engineering infrastructure. \\
\textbf{Arun Ahuja} contributed to agent and reward model development, imitation learning, reinforcement learning and engineering infrastructure. \\
\textbf{Federico Carnevale} contributed to agent and reward model development, imitation learning, reinforcement learning, running and analysis of experiments, writing and as a sub-effort lead. \\
\textbf{Petko Georgiev} contributed to environment development, data and tasks, running and analysis of experiments, engineering infrastructure, evaluation and as technical lead. \\
\textbf{Alex Goldin} contributed to project management. \\
\textbf{Alden Hung} contributed to agent and reward model development, imitation learning, reinforcement learning, running and analysis of experiments and as a sub-effort lead. \\
\textbf{Jessica Landon} contributed to data and tasks, evaluation, engineering infrastructure and as a sub-effort lead. \\
\textbf{Jirka Lhotka} contributed to environment development, running and analysis of experiments, engineer infrastructure, and writing. \\
\textbf{Timothy Lillicrap} contributed to environment development, agent and reward model development, imitation learning, reinforcement learning, data and tasks, evaluation, running and analysis of experiments, writing and as an effort lead. \\
\textbf{Alistair Muldal} contributed to data and tasks, evaluation, engineering infrastructure and as a sub-effort lead. \\
\textbf{George Powell} contributed to data and tasks, evaluation and engineering infrastructure. \\
\textbf{Adam Santoro} contributed to agent development, imitation learning, running and analysis of experiments and writing.\\
\textbf{Guy Scully}  contributed to project management. \\
\textbf{Sanjana Srivastava} contributed to reinforcement learning and running and analysis of experiments.
\textbf{Tamara von Glehn} contributed to environment development, agent and reward model development, imitation learning, reinforcement learning, running and analysis of experiments, engineering infrastructure and writing. \\
\textbf{Greg Wayne} contributed to environment development, agent and reward model development, imitation learning, reinforcement learning, data and tasks, evaluation, running and analysis of experiments, writing and as an effort lead. \\
\textbf{Nathaniel Wong} contributed to environment development. \\
\textbf{Chen Yan} contributed to agent and reward model development, imitation learning, reinforcement learning, running and analysis of experiments and writing. \\
\textbf{Rui Zhu} contributed to environment development, agent development, reinforcement learning, running and analysis of experiments and engineering infrastructure. \\ \\

\vspace{1mm}
\noindent
{\bf Corresponding Authors:} \\
Federico Carnevale (fedecarnev@deepmind.com), Greg Wayne (gregwayne@deepmind.com) \& Timothy Lillicrap (countzero@deepmind.com)

%% file: sections/acknowledgements.tex
\section{Acknowledgments}

The authors would like to thank Felix Hill, Daan Wierstra, Dario de Cesare, Mary Cassin, Arthur Brussee, Koray Kavukcuoglu and Matt Botvinick.

%% file: sections/appendix.tex
\section{Appendix} 

\subsection{Human-agent-human interactions (shared autonomy)} \label{sec:shared_autonomy}

In human-agent-human interactions (also known as shared autonomy), two players are paired with one agent. One of the humans assumes the role of setter, while the agent always assumes the role of solver. The second human (referred to as the co-pilot) can observe the interaction through the viewpoint of the agent and is able (and instructed) to take over control of the solver's avatar upon a prominent agent failure. After taking control, the co-pilot is instructed to complete the task and return control to the agent.

Human-agent-human interactions both probe the agent's skills and provide direct demonstrations of corrective behaviours where the agent is weakest. The instructions the setters used in the shared autonomy game match the type of instructions given in the human-agent game (Section \ref{sec:language_game_prompts}).

During a takeover period, the agent movement and language actions are suppressed, instead the avatar moves with the human controls as a human solver would in the human-human game. During that time, we make sure the agent is forced with the co-pilot's actions as its own previous actions to ensure consistency in the agent's internal representation through take-over periods. When recording shared autonomy episodes, in addition to the regular episode observation stream of images, actions and text, we also save a unique identifier representing whether the agent or human co-pilot is controlling the avatar. We use these indicators to identify candidate scenarios for the STS, as human interventions indicate behaviors where agents are making failures.

\subsection{Human feedback interface} \label{sec:interface}

\begin{figure}[t]
	\centering
	\includegraphics[width=0.7\textwidth]{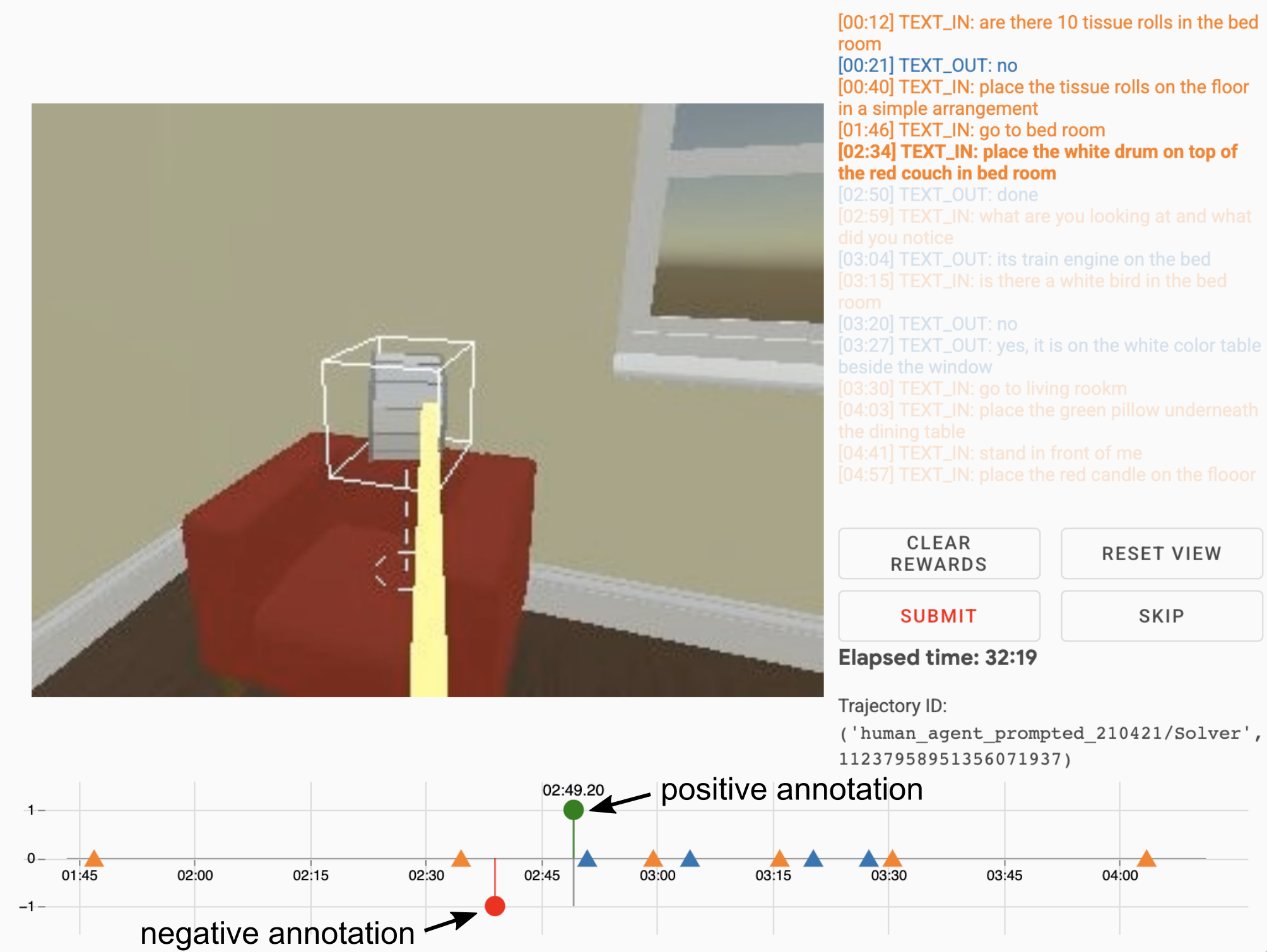}
	\caption{\textbf{Human feedback interface.} Raters browsed recorded episodes (including video and dialogue) and could move backwards and forwards through time, observing the interaction through the same view as the solver. Using this interface, they marked discrete moments during the episode where they considered that the solver made progress towards or regressed from the goal of the interaction. }
	\label{fig:interface}
\end{figure}

We created an interface where raters could browse recorded episodes of human setters interacting with human or agent solvers (Figure \ref{fig:interface}). Raters could scan through episodes by moving the mouse cursor left and right, observing the video through the same view as the solver toghether with the dialogue between two players. 

Raters were asked to mark discrete moments during the episode where the solver made progress towards or regressed from the goal of the interaction. Specifically, we ask them to mark a positive event when, ``(1) The solver completes an instruction successfully or (2) The solver does something that brings them closer to completing the instruction, and to mark a negative event when (1) The solver makes a clear mistake or (2) The solver does something that is not relevant to the instruction.'' 

We gave the following example:

\fbox{
\centering
\parbox{0.8\textwidth}{
If the instruction was to “put the orange duck on the shelf in the bedroom” then examples of “positive” events might be:

\begin{itemize}
    \item When the solver picks up the orange duck.
    \item When the solver moves into the bedroom.
    \item When the solver places the duck on the shelf.
\end{itemize}

Some examples of “negative” events might be:

\begin{itemize}
    \item If the solver picks up an object that isn't an orange duck.
    \item If the solver wanders into a different room that isn't the bedroom.
    \item If the solver places the orange duck somewhere that isn't on the shelf.
    \item If the solver says random words that aren't relevant to the instruction.
    \item If the solver knocks over other objects in the room.
\end{itemize}}}

For instructions that involve dialogue, we asked raters to annotate each of the solver’s utterances with a positive mark if it is correct and relevant to the instruction, and with a negative mark otherwise. For example:

\fbox{
\parbox{0.8\textwidth}{

Setter: Is there a purple rocket in the room? [In this case there is] \\
Solver: No	← Negative: incorrect answer \\
Solver: Yes  ← Positive: correct answer \\
\\
Setter: Put the ball in the box\\
Solver: The white or the red one?	← Positive: this is a relevant question\\
Setter: The red one \\
Solver: Three beds ← Negative: this has nothing to do with the  instruction
\\
}}

\subsection{Inter-temporal Bradley-Terry Model} \label{sec:ibt_method}

We describe abstract pseudocode for the IBT model in Algorithm \ref{alg:cap}. For convenience of notation, we write the algorithm with batch size 1, although in practice we used batches of size $96$.

\begin{algorithm}[H]
\caption{IBT training algorithm}\label{alg:cap}
\begin{algorithmic}
\For{training step k in (1, \dots, N)}
\State $x_{0:T}, y_{0:T} \sim \mathcal{D}$ \Comment{Sample observations and annotations of an episode from the dataset}
\For{ all time pairs $(t_1, t_2)$ in $(0, \dots, T)$ with $t_2 \geq t_1$}
\State $u_1 = U(x_{0:t_1})$ \Comment{Compute utility of trajectory up to $t_1$}
\State $u_2 = U(x_{0:t_2})$ \Comment{Compute utility of trajectory up to $t_2$}
\If{there are only positive annotations in $y_{t_1: t_2}$}
    \State $L = -\text{LogSigmoid}(u_2-u_1)$
\ElsIf{there are only negative annotations in $y_{t_1: t_2}$}
    \State $L = -\text{LogSigmoid}(u_1-u_2)$
\ElsIf{there are no annotations in $y_{t_1: t_2}$}
    \State $L = \text{C} \times (u_2-u_1)^2$ \Comment{C weighs the no annotation loss}
\EndIf
\State \text{Update parameters using gradient of loss} $L$
\EndFor
\EndFor
\end{algorithmic}
\end{algorithm}

\subsection{Architecture} \label{sec:architecture}

We used the same base architecture for the agent and the rewards model as \cite{iax2021b}. Briefly, a series of ResNet blocks downsample the incoming image, while language tokens index a learnable embedding table. Together these embeddings comprise the input to an 8-layer multi-modal Transformer, whose output is aggregated and provided as input to an LSTM memory. The output of the LSTM conditions both the hierarchical movement policy and language policy, implemented as an MLP (which produces 8 sets of consecutive movement actions) and a Transformer, respectively. Our base BC and BC+RL agents have 71M parameters. For the reward model we added an extra MLP head with a one-dimensional output representing the utility estimate $U$ at each time.

\subsection{Autoregressive Reward Model} \label{sec:autoregressive}

\begin{figure}[t]
	\centering
	\includegraphics[width=\textwidth]{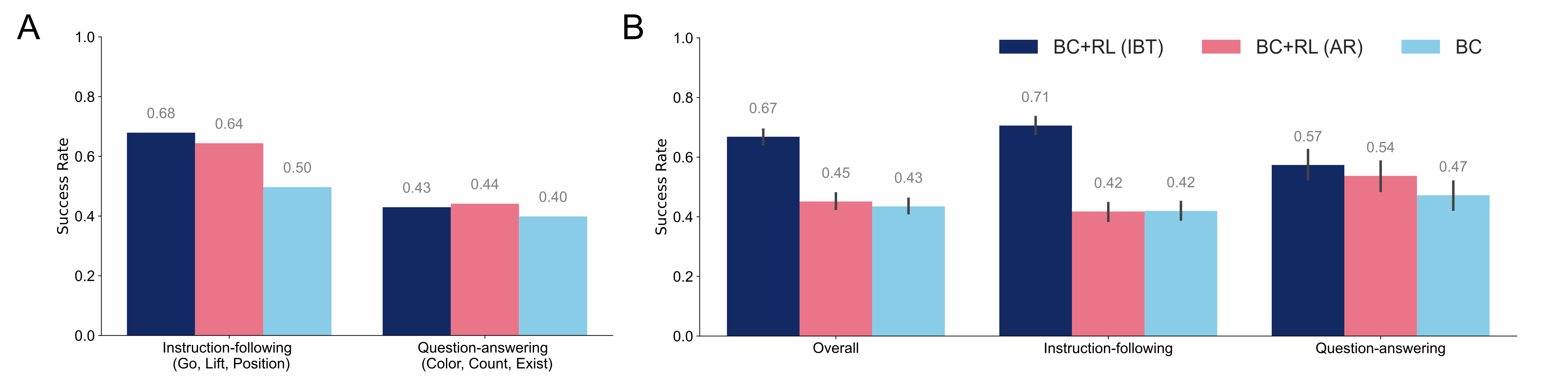}
	\caption{Autoregressive reward model. \textbf{A} Average success rate of BC+RL agents (IBT and AR) versus the BC baseline on the "Position" held out scripted task as a function of training steps, using IBT or autoregressive reward models. \textbf{B} Average success rate on STS scenarios.
}
	\label{fig:autoregressive}
\end{figure}

As an alternative to the IBT reward model described in Section \ref{sec:reward_model}, we also trained an autoregressive (AR) reward model to directly predict the human feedback at each timestep. Instead of an MLP utility head, this model has a head consisting of an LSTM followed by an autoregressive sequence model whose output is a three-way predictor of the probabilities of a positive, negative or no annotation for each timestep. During training, at each timestep the LSTM receives the output of the memory LSTM as well as the ground truth human annotation from the previous timestep (i.e., autoregressive conditioning), and the loss is the negative log-probability of the current annotation. When used as a reward model for training a BC+RL agent, the LSTM receives the output of the memory LSTM as well as the previous head output, and the classifier samples the annotation of $1$, $-1$ or $0$. These annotations are used directly as reward at each time-step. The same auxiliary losses, initialization and dataset augmentation as for the IBT reward model were used during reward model training. Figure \ref{fig:autoregressive} shows the performance of a BC+RL agent trained with this AR reward model on scripted tasks and STS scenarios. Although RL with the AR reward model outperformed the BC base agent, it was significantly worse than the agent trained with the IBT reward model.

\subsection{RL with pretrained setters}  \label{sec:pretrained_setters}

The setter-replay approach has important limitations. It does not easily permit learning back-and-forth interactions between players, and the set of interactions is limited to the data that we have collected. A promising approach to overcome these is to train setter agents: i.e., agents that can produce human-like instruction and questions to train solver agents. However, training setter agents is challenging; for setters to be useful, they need to generate instructions that are coherent, feasible in the environment, and diverse so that they generate good training data for solver agents.

Here we perform an initial exploration of training solver agents by interacting with learned setter agents. We first train setter agents via BC to imitate humans giving instructions or asking questions. Then we use these pre-trained (frozen) setters as part of the RL environment to train solver agents. The setter agent's architecture is exactly the same as the solver.

\begin{figure}[t]
	\centering
	\includegraphics[width=\textwidth]{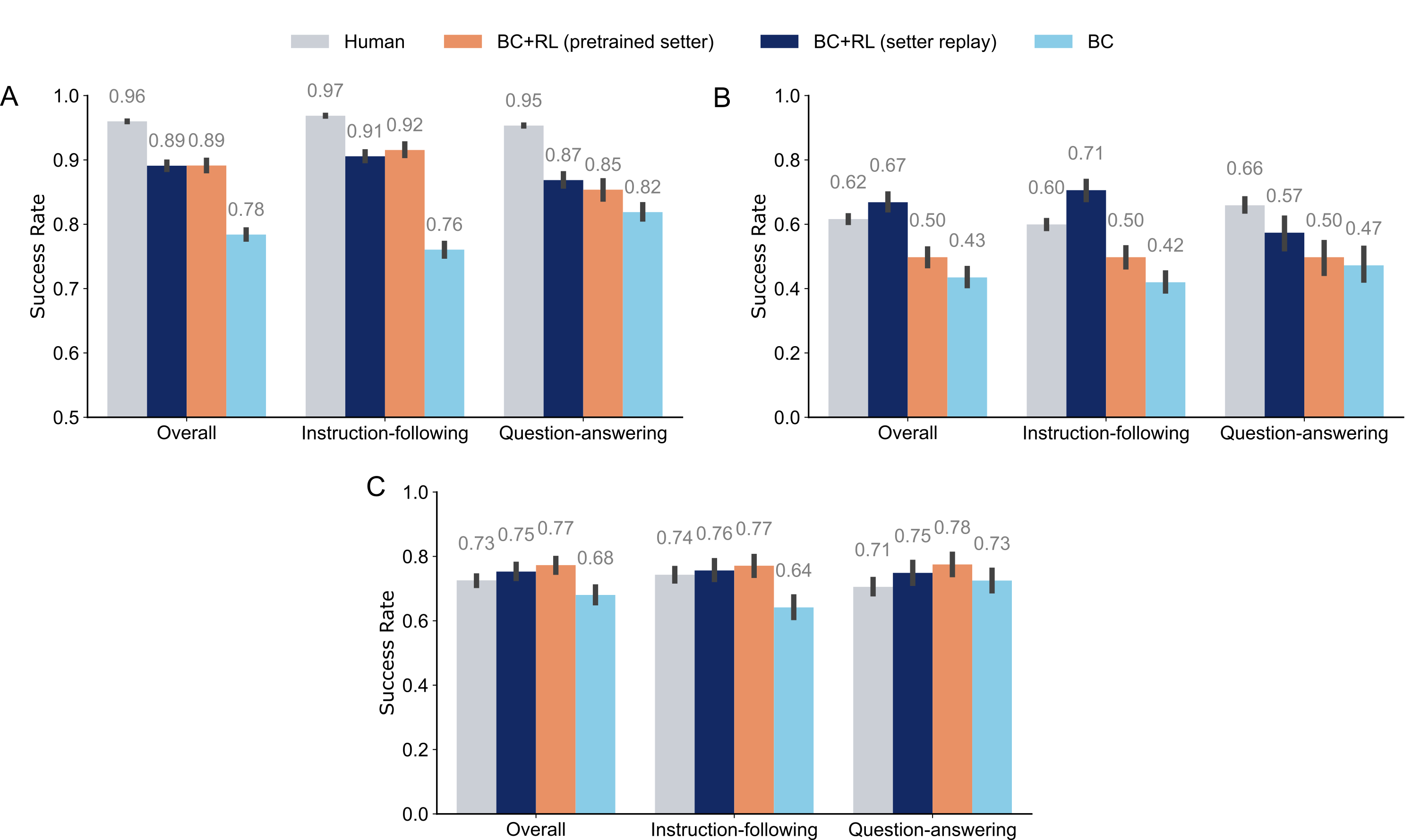}
	\caption{\textbf{RL with pre-trained setters}. \textbf{A} Interactive evaluation. Agents trained with pre-trained setters score similarly as agents trained using the setter-replay environment. \textbf{B} Standardised Test Suite evaluation. When evaluated on STS scenarios with challenging instructions, agents trained with pre-trained setters score worse. \textbf{C} We collected a second set of STS scenarios with easier instructions and, in this case, agents trained with pretrained setters score as well as those trained with setter-replay.
	\label{fig:pretrained_setter}}
\end{figure}

Figure \ref{fig:pretrained_setter} shows the evaluation of solver agents trained with pre-trained setters (PS agents) compared to agents trained using the setter-replay environment (SR agents). PS agents are as good as SR agents on interactive evaluation (Figure \ref{fig:pretrained_setter}A). However, when evaluated on STS, they performed significantly worse (Figure \ref{fig:pretrained_setter}B). We hypothesised that the most challenging STS scenarios were the differentiator. To test this, we evaluated agents on the original set of scenarios used in \cite{iax2022}, which does not use episodes with challenging instructions. In this case, PS agents perform as well as SR agents (Figure \ref{fig:pretrained_setter}C). Given these experiments, we feel that improving our understanding of methods to train agents in this kind of ``self-play'' is an important line of future work. Videos of setter and solver agents interacting with each other may be found at \url{https://youtu.be/pCloAOyZZxU}.

\subsection{Training setup and hyperparameters} \label{sec:hypers}

We trained agents with a combination of supervised learning-based behavioural cloning to human interactions (BC), contrastive self-supervised learning (CSS) and reinforcement learning (RL). Table \ref{table:hyper} contains a list of all training hyperparameters. 

We adopted a distributed training framework where agent trajectories were generated on “actor” computers and then sent to a “learner” in a [T, B] format, where T is the sequence length and B the batch size. Actors were split into \textit{dataset} actors for behavioural cloning and self-supervised learning, and \textit{setter-replay} actors for reinforcement learning. Batches of trajectories from each type of actor were combined, making a full batch of size 2 × B, with different losses applied to each. 

For behavioural cloning, we used the same setup as in \cite{iax2021b}. Observations from human-human episodes were sequentially given to the agent to obtain movement and language actions distributions for each timestep and we maximized the likelihood of the ground truth actions taken by humans. We used different coefficient for movement and language actions ($w^{\text{BC}}_{\text{move}} = 1$ and $w^{\text{BC}}_{\text{lang}} = 5$, respectively). 

For contrastive self-supervised learning, we used the same technique as in \cite{iax2021a, iax2021b} wherein the agent must predict whether vision and language embeddings match (i.e., they are produced from a trajectory from the dataset as normal), or they do not match (i.e., visual embeddings are produced from the input image of one trajectory in the dataset, and language embeddings are produced from the language input from a different trajectory). We implemented this by adding an MLP discriminator that produces a binary prediction and used the same batch of data used for the behavioural cloning for the “matches”, and a shuffling of vision and language embeddings for the mis-matches. We added this auxiliary loss to the total loss with a weight $w^{\text{CSS}}=1$.

For reinforcement learning, we used V-trace \citep{espeholt2018}. The value function baseline was implemented in the agent by an additional MLP head with a hidden layer size of 512 taking in the same inputs as policy heads do. Both the movement and language policy shared the same rewards and value function. We used a discount factor of $\gamma=0.96$ and different coefficients for weighing the movement policy updates ($w^{\text{RL}}_{\text{move}}=0.5$), language policy updates ($w^{\text{RL}}_{\text{lang}}=0.1$) and the value function loss ($w^{\text{V}}=1$). 

In the RL experiments, we initialize the agent's parameters with those of a pretrained BC agent and continue training with the BC, CSS and RL objectives. Additionally, we add a loss that penalises the KL divergence between the training policy and the initial BC policy. In practice, we found that the only component of the policy that required this penalty was the one controlling when to produce language outputs, for which we used a weight of $w^{\text{KL}}_{\text{lang}}=0.001$. 

For all experiments we used the ADAM optimizer \citep{kingma2014} with $\beta_1=0.$, $\beta_2=0.999$ and learning rate $\lambda=1e-4$.

\begin{table}[!htb]
\centering
        \begin{tabular}{p{1cm}<{\raggedright} p{1cm} p{4cm}<{\raggedright}}
          \toprule
                       & \textbf{Values}     & \textbf{Description} \\ 
          \midrule
            T                                   & 50        & Sequence length  \\
            B                                   & 96        & Batch size  \\
            $w^{\text{BC}}_{\text{move}}$       & 1.        & BC language weight \\
            $w^{\text{BC}}_{\text{lang}}$       & 5.        & BC language weight \\
            $w^{\text{CSS}}$                    & 1.        & CSS weight \\
            $w^{\text{RL}}_{\text{move}}$       & 0.5       & RL movement weight\\
            $w^{\text{RL}}_{\text{lang}}$       & 0.1       & RL language weight    \\
            $w^{\text{V}}$                      & 1.        & Value weight    \\
            $w^{\text{KL}}_{\text{lang}}$       & 1e-3      & no-op KL weight\\
            $\gamma$                            & 0.96      & Discount Factor \\
            $\beta_1$                           & 0.        & ADAM beta 1  \\
            $\beta_2$                           & 0.999     & ADAM beta 2 \\
            $\lambda$                           & 1e-4      & Learning Rate \\
          \bottomrule
        \end{tabular}
        \captionsetup{justification=centering}
        \caption{BC+RL agent hyperparameters.}
        \label{table:hyper} 
\end{table}

\subsection{Language games prompts used for human-agent interactions} \label{sec:language_game_prompts}

When collecting human-agent interactions we gave the human setter the following prompts to indicate what kind of instruction or question they should pose to the solver. The prompt consisted of a textual cue indicating the general type of instruction, combined with a modifier that stipulated additional constraints that the setter's instruction must satisfy.

\subsubsection*{Instruction-following}
Ask the other player to touch an object using another object. \\
Ask the other player to perform an activity of your choice. \\
Ask the other player to hand you something, which you then hold. \\
Ask the other player to stand in some position relative to you. \\
Ask the other player to move a group of objects into a simple arrangement. \\
Ask the other player to put something on top of something else. \\
Ask the other player to put something underneath something else. \\
Ask the other player to lift something. \\
Ask the other player to do something, then when they are finished, ask them to do something else. \\
Ask the other player to reposition something in the room. \\
Ask the other player to bring you one or more objects. \\
Ask the other player to push an object around using another object. \\
Ask the other player to position something relative to something else. \\
Ask the other player to put three or more specific objects in a row. \\
Ask the other player to go somewhere.

\subsubsection*{Question-answering}
Ask the other player a yes/no question about something in the room. \\
Ask the other player to describe where something is. \\
Ask the other player to count something. \\
Ask the other player to say what they are looking at or noticing right now. \\
Ask a question about the color of something. \\
Ask the other player whether two things have the same or different colors. \\
Ask the other player a yes/no question about numbers of objects in the room. \\
Ask the other player whether a particular thing exists in the room. \\
Ask the other player any question about the room. \\
Ask the other player whether one object is larger, smaller, or the same size as another object.

\subsubsection*{Modifiers}
Refer to objects by colour. \\
Refer to location by colour. \\
Refer to objects by location. \\
Use shape words. Try to use shape words like: circular, rectangular, round,
pointy, long. \\
Use proximity words. Try to use words like: near, far, close to, next to. \\
Use horizontal position words. Try to use words like: in front, behind, left of, right of,
between. \\
Use vertical position words. Try to use words like: on top, beneath, above, below. \\
Use negation words. Try to use words like: not, isn't. \\
Use quantifier words. Try to use words like: some, all, most, many, none. \\
Do not use the words: bed, door, window.

\pagebreak
\section{STS scenarios} \label{sec:sts_scenarios}

Below is the list of setter instructions from the 162 scenarios used for STS evaluation. The text is taken verbatim from what humans asked the agents in the original episodes. 

\begin{multicols}{2}
\tiny
\noindent
0. Add the circular objects into the green bin in play room\\
1. Is bed larger,smaller or same size as sofa in bed room?\\
2. Build a tower with white objects in the living room\\
3. Place the chairs on the floor in a row in the living room\\
4. What is color of chest which is near me\\
5. Build tower with tissue rolls\\
6. Lift boat in pantry\\
7. Bring me one or more objects from bed room?\\
8. Built a tower using soap,book and tissue roil which is in restroom\\
9. Form a triangle with white objects near to white furniture in the living room\\
10. What are you noticing at right now?\\
11. Add any 2 same color objects into cyan bin in bedroom\\
12. Put the train engine on the wardrode?\\
13. Put any three objects in a row\\
14. Bring me two tissue rolls\\
15. Take books and form a triangle\\
16. Make a single row uisng small bjects in living room.\\
17. To go dining table in living room\\
18. Lift the red color object which flies in the air in bedroom\\
19. Count bananas in pantry room\\
20. Hit the yellow toy which is not near to the table in living room.\\
21. Arrange a  row  with the white color objects on the floor in the living area \\
22. Count the number of objects which flys in the sky in the bed room?\\
23. Remove all objects from the box which is in green color in play room\\
24. Arrange 3 objects in ascending order rule in living room\\
25. Hand me violet candle in living area\\
26. Bring me eatables from pantry room.\\
27. Touch the rack using ball?\\
28. Place a pillow on the bed, in bed room \\
29. Place the mug under the table of living room.\\
30. Place orange color cushion under red color armchair which is in bedroom\\
31. Where is the pink ball?\\
32. Touch the potted plant using the bird toy\\
33. Push the box with shoe in bedroom\\
34. Count the pointy objects in the living room\\
35. Push red color object with yellow color object which are in the pantry .\\
36. Build a tower with books in living area\\
37. Touch car toy with rocket which are on the floor of bed room\\
38. Place the bottle on the table in living romm/\\
39. Hand me a animal toy from bed room\\
40. Place the flying object underneath the table?\\
41. Touch the blue tissue roll using the robot in the restroom\\
42. Make a row using white colour objects in the living room\\
43. Where is the orange duck in the wash room\\
44. Place soaps underneath the ledge in bathroom\\
45. Compare table and rack and tell which one is larger, smaller or they of same size\\
46. How many white colour objects are there n living room\\
47. Is there five green color objects were present in the bed room or not \\
48. Place the object which is used to dry  next to sitting object \\
49. Describe where is the small white object in playroom?\\
50. Lift the basket ball in the bed room\\
51. Stand far from me\\
52. Bring me 2 water swimming objects \\
53. Push the potted plant using the plate?\\
54. Where is yellow duck present in house\\
55. Form a row with green color objects in living room\\
56. Put the bin on top of dining table \\
57. Hand me a cushion from bedroom which is right of bed.\\
58. Lift red color round object which is on top of the table in living room\\
59. Whether 2 drum have same color or different color in purple room?\\
60. Does there is the red object between the ball and candle in the living room?\\
61. Build a tower with three different colored and same type of rectangular objects.\\
62. Throw white teddy from the playroom\\
63. Arrange four tissues in a row \\
64. Hand over the plate?\\
65. Where is the bed?\\
66. Move the eatable objects into a row \\
67. Lift the purple item in the pantry\\
68. Place the bus toy on the ledge of living room.\\
69. What is the color of the footrest in living room?\\
70. Remove all objects from the bin in play room\\
71. Break the plate on the table?\\
72. Push the train engine with water bird\\
73. Put the red pillow underneath the bin in bed room\\
74. Form a circle of objects in pantry\\
75. Remove the objects from the bin box in bed room?\\
76. Touch the red lamp with the blue one \\
77. Touch the white object which is next to the wardrobe with the cyan colour object in the living room?\\
78. Hit the book using another book\\
79. Place the siting object on table in this room \\
80. Bring a mirror frame for me\\
81. Build a tower in the rest room?\\
82. Place the one circular. one pointy, one rectangular object in a row on floor in living room.\\
83. Name the objects which are in pantry room \\
84. Tell me the count of candles \\
85. Count the objects which are in red rack\\
86. Eat two fruits which are in pantry room\\
87. Place three white objects in a row in playroom?\\
88. Move three rectangular objects into a row on the floor in living room\\
89. Is there a musical instrument in play room\\
90. Tell me any three objects on table\\
91. Place the yellow balls in the white bin\\
92. Where is the ball in playroom which is not used for the tennis?\\
93. Bring me 2 tissue rolls from pantry to living room\\
94. Place the grapes on the dining table and then place the lemon on the dining table in the living room\\
95. Form a triangle with blocks in play room\\
96. Add two books in the box\\
97. Where is the red color lamp in this room \\
98. Tell me the names of playing objects in living room.\\
99. Place olive pillow on sofa in living room'\\
100. Give me the count of pillows which are on  the floor of the room which has bed.\\
101. Make a circle with the objects on the floor in the play area \\
102. What is the color of the object in the wall frame near the arm chair in living room\\
103. Stand in between two chest which are in front of me\\
104. Lift the mini chair in playroom\\
105. What is the color of rack?\\
106. Put objects on bed which are on the floor in bedroom\\
107. Bring me two objects from bedroom.\\
108. Count the number of tissue rolls?\\
109. Where is grey pillow?\\
110. Make a row on floor with 3 books and 3 potted plants in the living room\\
111. Hand me a teddy  which is in play room.\\
112. Take 4 books and form a square in living room\\
113. Form a circle with the few items in the living room\\
114. Push peach tissue with blue book in rest room.\\
115. Bring two fruits from pantry\\
116. Throw the mirror out of the playroom.\\
117. Make a row with 3 different colour objects in living room\\
118. Arrange 4 pointy objects in a square shape in the bed room\\
119. Where is green color flying object in living room \\
120. Remove the rectangular object from box  \\
121. Lift the drum and place on the blue pillow?\\
122. Tell me the names of pointy objects in this room ?\\
123. The rack and wardrobe in living room have same color or different\\
124. Hand me hair drier in bed room.\\
125. Go and stand opposit eto me\\
126. Count the object in the rack.\\
127. Put the white book on red chair in living room \\
128. Put the paper roll in the bathtub and then lift the soap bar in restroom.\\
129. Hit the vehicle with the animal\\
130. Make a triangle by using 2 square objects and 1 rectangular object in the same room as bed exists in the room\\
131. Form a triangle by using 3 books in the living room\\
132. Put the light object under the dining furniture in the living area\\
133. Place three pillows in a row on floor\\
134. Arrange four objects in a line on floor\\
135. Lift the object which is at right side column at the top in the rack in the living room.\\
136. Put three red objects in a row on the floor small to big?\\
137. Place a cyan pointy object on the bed\\
138. Form a circle with the objects in the bedroom\\
139. Build a tower with four different objects in living room.\\
140. Put three objects on top of bath tub plank n a row \\
141. What is color of rack in living room\\
142. Form a triangle shape on bedroom floor by using cushions\\
143. Bring two plants from living room\\
144. Describe where is yellow rack"\\
145. Hit the candle using the pillow which is left of airplane in the living room \\
146. What is the count of books in rack in living room?\\
147. Tell me names of objects in pantry\\
148. What is the color of rack in bed room ..\\
149. Add the train toy into the green bin?\\
150. Place the stool underneath the dining table in the living room\\
151. Put the object we pore water to it underneath of dining table in living room\\
152. Hand me the pear fruit in the pantry room \\
153. Push the chair using the plotted plant\\
154. Find white colored robot in this house and go near to it\\
155. Place the arm chair far to bed in bed room\\
156. Add tennis ball into bin in play room\\
157. Make a row of 3 boats in bedroom\\
158. Where is the purple color object in living room\\
159. Wether mug exists in living room\\
160. Push the pillow in living room\\
161. Bring me 3 cushions\\
162. Put all objects underneath ledge in pantry\\
\end{multicols}